\documentclass{article}
\usepackage{spconf,amsmath,graphicx, mathtools}
\usepackage{multirow,amssymb,hyperref}
\usepackage{booktabs,url}
\usepackage{empheq}
\usepackage{bm}
\usepackage{subfiles}

\usepackage{mathtools}

\DeclarePairedDelimiterX{\infdivx}[2]{(}{)}{%
	#1\;\delimsize\|\;#2%
}
\newcommand{\KL}{{\tt KL}\infdivx}

\newcommand{\Dir}{{\tt Dir}}
\newcommand{\GLEU}{{\tt GLEU}}
\newcommand{\AUC}{{\tt AUC}}
\newcommand{\AUCRR}{{\tt AUC_{RR}}}
\newcommand{\REFNLL}{\ref{eq: nll distdist}}
\newcommand{\REFKL}{\ref{eq: kl distdist}}

\DeclareMathOperator*{\argmax}{arg\,max}

\title{Ensemble Distillation Approaches for Grammatical Error Correction}
\twoauthors
{Y. Fathullah, M.J.F. Gales\sthanks{This research was partly funded by  Cambridge Assessment, who also gave permission to use the BULATS data.}}
{Cambridge University Engineering Department \\ \small{\texttt{\{yf286,mjfg\}@eng.cam.ac.uk}}}
{A. Malinin}
{Yandex, Moscow, Russia \\ \small{\texttt{am969@yandex-team.ru}}}
\begin{document}
	\ninept
	
	\maketitle
	
	\begin{abstract}
		Ensemble approaches are commonly used techniques to improving a system by combining multiple model predictions. Additionally these schemes allow the uncertainty, as well as the source of the uncertainty, to be derived for the prediction. Unfortunately these benefits come at a computational and memory cost. To address this problem ensemble distillation (EnD) and more recently ensemble distribution distillation (EnDD) have been proposed that compress the ensemble into a single model, representing either the ensemble average prediction or prediction distribution respectively. This paper examines the application of both these distillation approaches to a sequence prediction task, grammatical error correction (GEC). This is an important application area for language learning tasks as it can yield highly useful feedback to the learner. It is, however, more challenging than the standard tasks investigated for distillation as the prediction of any grammatical correction to a word will be highly dependent on both the input sequence and the generated output history for the word.  The performance of both EnD and EnDD are evaluated on both publicly available GEC tasks as well as a spoken language task. 
		
	\end{abstract}
	
	
	\section{Introduction}
	\label{sec:intro}
	
	Knowledge distillation (KD) is a general approach in machine learning where predictions from a more complex model are modelled by a simpler model, or an ensemble of models represented by a single model \cite{knowledge distillation}. The aim is to reduce the computational cost and memory requirements for deployment. It has been applied successfully in many domains in deep learning, such as object detection \cite{dist object detection}, natural language processing \cite{nlp kd 1, nlp kd 2}, acoustic models \cite{DBLP:conf/interspeech/WongG16,dist lm}, and also in adversarial defence \cite{end for adverserial attack}. This paper focuses on applying ensemble distillation approaches for sequence to sequence tasks, in this case grammatical error correction (GEC).
	
	Deep ensembles generally perform better than single models, with the added benefit of providing  measures of prediction uncertainty  \cite{murphy, seed ensemble}. Unfortunately these benefits come at the cost of higher computational power and memory. To address this problem, ensembles can be distilled into a single model. Normally, distillation results in information about the uncertainty in the distilled model's prediction being lost \cite{prior net}. One approach to maintaining these uncertainty measures after ensemble distillation is \textit{Ensemble Distribution Distillation} \cite{ensemble dist dist}. Here the distilled student models the distribution of the categorical predictions from the ensemble, allowing it to retain more information about its teacher ensemble. Usually a Dirichlet distribution is used for the distribution over categorical distributions. This model can be challenging to train as the student model is required to represent significantly more information from the teacher ensemble than the standard, cross-entropy based, ensemble distillation \cite{ensemble dist dist}. If the distilled model is well trained, however, it should efficiently yield good results and enable, for example, detecting whether inputs are in or out-of-distribution, and yield insight into why a prediction is (un)reliable \cite{ensemble dist dist, prior net, ensemble uncertanity spoken}.
	
	This paper examines distillation approaches for sequence to sequence tasks such as machine translation and grammatical error correction. For these tasks the inference computational cost is significantly larger than "static" tasks, as the decoding process must combine the sequence predictions from each individual in the ensemble \cite{DBLP:conf/interspeech/WongG16,seq endd}.  Distillation schemes already exist which tackle the problem of reducing an ensemble, to save resources while also maintaining performance. Distribution distillation methods however, have not been explored for sequence models \cite{structured prediction, understanding kd}. This distribution distillation would enable efficient uncertainty measures for tasks such as speech-to-text systems and neural machine translation. The application area explored here is GEC \cite{gec dnn task}. The aim is to extend previous work on uncertainty for spoken language assessment \cite{ensemble uncertanity spoken, gec dnn task} to the problem of providing feedback to the learner, and one challenge for general GEC systems is the difference between acceptable grammar for written and spoken English.
	
	Initially, standard distillation approach for sequence-to-sequence models, in particular auto-regressive sequence-to-sequence, are described. Note rather than modelling the (approximate) sequence posterior as in \cite{DBLP:conf/interspeech/WongG16,seq endd}, this work models the token-level (word) posterior as this enables a simple extension to distribution distillation for this form of model, and can be efficiently used to derive uncertainty measures~\cite{structured prediction}. The challenges of applying distribution distillation are then discussed, building on top of the work done in \cite{ensemble dist dist}. Two possible optimisation approaches for ensemble distribution distillation are described, as well combining standard distillation and distribution distillation models together. These approaches are all evaluated on the task of performing grammatical error correction, on both written data, where the training and test data are from the same domain, and speech data where there is a mismatch. Uncertainty performance is assessed on these tasks, and compared to rejection based on the manual, reference, corrections. 
	
	
	\section{Ensemble Distribution Distillation}
	\label{sec:ens dist dist}
	
	Knowledge and distribution distillation are motivated by the need for high performance systems with low memory, power, and time, requirements, whilst also yielding uncertainty measures \cite{prior net, ensemble dist dist, seq endd, structured prediction, compression e2e}. There are many ways in which this can be achieved: here we will present work on knowledge distillation and distribution distillation.
	In this section, we will focus on static models, which are used for image classification \cite{prior net} and spoken language assessment \cite{ensemble uncertanity spoken}. A teacher ensemble is represented by a set of parameters $\{\bm\theta^{(m)}\}_{m=1}^{M}$, and the parameters of the standard and distribution distilled students will be represented by $\bm\phi$ and $\bm\lambda$, respectively (to signify a fundamental difference). Assuming that the teacher ensemble are drawn from the posterior ${\tt P}(\bm\theta|\mathcal{D})$, then:
	\begin{align*}
	{\tt P}(y|\bm x; \bm\theta^{(m)}) = {\tt P}(y|\bm{\pi}^{(m)}), \medspace\bm{\pi}^{(m)} = \bm{f}(\bm x; \bm\theta^{(m)})
	\end{align*}
	where $\pi_i$ represents the probability ${\tt P}(y = \omega_i)$ and $\bm x$ is the input. The set of categoricals $\{\bm{\pi}^{(m)}\}_{m=1}^{M}$ can then be used to train new students. 
	In standard distillation \cite{knowledge distillation}, a student $\bm\phi$ is trained to emulate the predictive distribution by minimising:
	\begin{align*}
	\mathcal{L}(\bm{\phi}) & = \KL[\bigg]{\left(\frac{1}{M}\sum_{m=1}^M\bm{\pi}^{(m)}\right)}{{\tt P}( y |\bm{x}; \bm{\phi})}
	\end{align*}
	This method essentially aims to let the student predict the mean of the teacher while information about the spread (or disagreement) between members of the ensemble is lost. This means that although distillation can significantly improve memory and power consumption, the uncertainty metrics will be lost.
	
	Instead, following \cite{ensemble dist dist}, we let a new student model $\bm\lambda$ predict the parameters $\bm\alpha$ of a Dirichlet distribution $\Dir(\cdot\medspace;\bm\alpha)$:
	\begin{align*}
	{\tt P}(\bm\pi|\bm x, \bm\lambda) = \Dir(\bm\pi;\bm\alpha), \medspace \bm\alpha = \bm{f}(\bm x; \bm\lambda), \medspace \alpha_{0} = \bm{1}^T\bm\alpha
	\end{align*}
	The Dirichlet is a distribution over categoricals, and can encapsulate information about the uncertainty and spread in $\{\bm{\pi}^{(m)}\}_{m=1}^{M}$. The training can then be based on negative log-likelihood (NLL):
	\begin{align*}
	\mathcal{L}(\bm\lambda) = -\frac{1}{M} \sum_{m = 1}^{M} \ln\bigg(\Dir(\bm{\pi}^{(m)};\bm\alpha)\bigg)
	\end{align*}
	Now it is possible to quantify uncertainty in the prediction. Three measures of uncertainty, and their associated source, are used in this work: \textit{total}, \textit{expected data} and \textit{knowledge} \textit{uncertainty}. These are  related by~\cite{back BNN,dropout ensemble}:
	\begin{align}\label{eq: static distdist unc}
	\underbrace{ \mathcal{I}[y,\bm\pi| \bm{x}, \bm\lambda] }_{\substack{Knowledge \\ Uncertainty}}
	= \underbrace{\mathcal{H}\big[{\tt P}(y|\bm x; \bm\lambda)\big]}_{\substack{Total \\ Uncertainty}}
	- \underbrace{\mathbb{E}_{{\tt P}(\bm\pi|\bm x; \bm\lambda)}\big[\mathcal{H}[{\tt P}(y|\bm\pi)]\big]}_{\substack{Expected\ Data \\ Uncertainty}}
	\end{align}
	These uncertainties have different properties depending on whether the input, $\bm{x}$, is in the same or different domain (distribution) to the training set. If $\bm{x}$ is in-domain (ID), the ensemble members should return consistent predictions, giving low expected data uncertainty (DU). If $\bm{x}$ is out-of-domain (OOD), the members will generate inconsistent predictions giving high total uncertainty (TU) \cite{adverserial uncertainty, ygal uncertainty, dataset shift}. 
	
	\section{Sequence Ensemble Distillation}
	\label{sec:seq dist dist}
	
	Applying both standard and distribution distillation to sequence models adds another layer of complexity. This section covers token-level distillation schemes, as these allow uncertainties for individual tokens, words, as well as being combined to yield sequence-level uncertainties \cite{structured prediction}. Extending the notation in the previous section to sequence-to-sequence models, the pair $(\bm x,\bm y)$ denotes the input-output reference sequence pair, and when necessary, $\bm{\hat y}$ represents the corresponding predicted sequence for ${\bm x}$. The teacher ensemble $\{\bm\theta^{(m)}\}_{m=1}^M$ now makes predictions by taking expectations over the following ensemble member predictions\footnote{For a discussion of alternative approaches to ensemble predictions based on sequences see~\cite{structured prediction}.}:
	\begin{align*}
	{\tt P}(y_l|\bm{y}_{<l}, \bm x; \bm\theta^{(m)}) = {\tt P}(y_l|\bm{\pi}_l^{(m)}), \medspace\bm{\pi}_l^{(m)} = \bm{f}(\bm{y}_{<l}, \bm x; \bm\theta^{(m)})
	\end{align*}
	and students are instead trained on the set $\{\bm\pi^{(1:M)}_l\}_{l = 1}^{L}$.
	
	\subsection{Ensemble Distillation}
	\label{ssec: end}
	
	For sequence models, distillation can be performed in multiple ways \cite{seq endd, more seq kd}. The approach adopted in this work is \textit{token-level knowledge distillation}, and is one of the simplest methods. The teacher and student $\bm\phi$ use the same reference back-history ${\bm y}_{<l}$ (teacher-forcing) and input $\bm x$. The KL-divergence between the ensemble and student token-level categorical distributions is then minimised:
	\begin{align}\label{eq:seq dist}
	\mathcal{L}(\bm\phi)=\frac{1}{L}\sum_{l = 1}^{L}  \KL[\bigg]{\left(\frac{1}{M}\sum_{m=1}^M{\bm{\pi}}_l^{(m)}\right)}{{\tt P}\big( y_l |\bm{y}_{<l}, \bm{x}; \bm{\phi}\big)}
	\end{align}
	
	Extending distribution distillation described in the previous section, this section introduces distribution distillation for sequence models. For token-level  distribution distillation, the student with parameters $\bm\lambda$ predicts the Dirichlet distribution with parameters $\bm\alpha_l$ for the $l$-th token: 
	\begin{align*}
	{\tt P}(\bm\pi|\bm{y}_{<l}, \bm x; \bm\lambda) = \Dir(\bm\pi;\bm\alpha_l), \medspace \bm\alpha_l = \bm{f}(\bm{y}_{<l}, \bm x; \bm\lambda)
	\end{align*}
	Given $\{\bm\pi^{(1:M)}_l\}_{l = 1}^{L}$ from the sequence ensemble, the distribution distilled model can be trained using negative log-likelihood (NLL):
	\begin{align}\label{eq: nll distdist}
	\mathcal{L}(\bm\lambda) = -\frac{1}{ML} \sum_{l = 1}^{L} \sum_{m = 1}^{M} \ln\bigg(\Dir(\bm{\pi}_l^{(m)};\bm\alpha_l)\bigg)
	\end{align}
	Eq. (\ref{eq: nll distdist}) optimizes the parameters of the distribution distilled model directly from the ensemble predictions for each time instance. 
	Though this yields an appropriate criterion to find ${\bm\lambda}$, predicting the distribution over the ensemble members for all back-histories, it may be very challenging to optimise the network parameters. To simplify this optimization, a two stage approach may be adopted. First, the predictions of the ensemble for each back-history is modelled by a Dirichlet distribution with parameters ${\tilde{\bm\alpha}}_l$ where 
	\begin{align*}
	\bm{\tilde\alpha}_l = \argmax_{\bm\alpha} \bigg\{ \frac{1}{M} \sum_{m = 1}^{M} \ln\Big(\Dir(\bm{\pi}_l^{(m)};\bm\alpha)\Big) \bigg\}
	\end{align*}
	The distribution distillation parameters are now trained to minimize the KL-divergence between this ensemble Dirichlet and the predicted distilled Dirichlet:
	\begin{align}\label{eq: kl distdist}
	\mathcal{L}(\bm\lambda) = \frac{1}{L} \sum_{l = 1}^{L} \KL[\Big]{\Dir(\bm\pi;\bm{\tilde\alpha}_l)}{\Dir(\bm\pi;\bm{\alpha}_l)}
	\end{align}
	This has the same general form as eq. (\ref{eq:seq dist}) but is based on the KL-divergence between distributions of distributions, rather than just the expected posterior distribution based on the ensemble. 
	
	Once a distribution distilled sequence model has been obtained, the probability of predicting class $c$ is then:
	\begin{align*}
	{\tt P}(y = \omega_c |\bm{y}_{<l}, \bm{x}; \bm{\lambda}) = 
	\mathbb{E}_{{\tt P}(\bm\pi|\bm{y}_{<l}, \bm{x}; \bm{\lambda})}\Big[{\tt P}(y = \omega_c |\bm\pi)\Big] = \frac{\alpha_{l, c}}{\alpha_{l, 0}}
	\end{align*}
	as would be expected when the student parametrises a Dirichlet. This shows that achieving sequence based distribution distillation is a straightforward generalization of both ensemble distribution and sequence distillation.
	
	Both eq. (\ref{eq: nll distdist}) and (\ref{eq: kl distdist}) result in viable objectives for obtaining a distribution distilled sequence model $\bm\lambda$, and from which uncertainty metrics can be derived. Instead of calculating entropy over the classes as  in eq. (\ref{eq: static distdist unc}), uncertainties for sequence models have to enumerate all possible sequences $\bm y \in \mathcal{Y}$. Let $\bm\Pi = \{\bm\pi_l\}_{l=1}^{L}$ be the sequence of categorical distributions from which the sequence $\bm y$ is generated:
	\begin{align*}
	{\tt P}(\bm\Pi|\bm x; \bm\lambda) = \prod_{l = 1}^{L}\Dir(\bm\pi_l;\bm\alpha_l), \medspace {\tt P}(\bm y|\bm\Pi) = \prod_{l = 1}^{L} {\tt P}(y_l|\bm\pi_l)
	\end{align*}
	The sequence uncertainties for the student $\bm\lambda$ can be expressed as:
	\begin{align}\label{eq: seq distdist unc}
	\underbrace{ \mathcal{I}[\bm y,\bm \Pi| \bm{x}; \bm\lambda] }_{\substack{Knowledge \\ Uncertainty}}
	= \underbrace{\mathcal{H}\big[{\tt P}(\bm y|\bm x; \bm \lambda)\big]}_{\substack{Total \\ Uncertainty}}
	- \underbrace{\mathbb{E}_{{\tt P}(\bm\Pi|\bm x; \bm\lambda)}\big[\mathcal{H}[{\tt P}(\bm y|\bm\Pi)]\big]}_{\substack{Expected\ Data \\ Uncertainty}}
	\end{align}
	However, as noted in \cite{structured prediction}, calculating the uncertainties require the intractable computation of enumerating all $\bm y \in \mathcal{Y}$. Instead the same approximations made in \cite{structured prediction} will be utilized here. Given $S$ sampled sequences: $\forall \bm{y}_{<l}^{(s)} \subset \bm{y}^{(s)} \sim {\tt P}(\bm{y}|\bm{x}, \bm\lambda)$, the following Monte-Carlo approximations can be made:
	\begingroup\makeatletter\def\f@size{8.8}\check@mathfonts
	\def\maketag@@@#1{\hbox{\m@th\small\normalfont#1}}%
	\begin{align} 
	\label{eq:total approx}
	\mathcal{H}\big[{\tt P}(\bm{y}|\bm{x}; \bm\lambda)\big]
	\approx &
	\frac{1}{S} \sum_{s, l} \mathcal{H}\big[{\tt P}(y_l |\bm{y}_{<l}^{(s)}, \bm{x}; \bm\lambda)\big] \\
	\label{eq:data approx}
	\mathbb{E}_{{\tt P}(\bm\Pi|\bm x; \bm\lambda)}\big[\mathcal{H}[{\tt P}(\bm y|\bm\Pi)]\big]
	\approx &
	\frac{1}{S} \sum_{s, l} \mathbb{E}_{{\tt P}(\bm\pi|\bm{y}_{<l}^{(s)}, \bm x; \bm\lambda)}\big[\mathcal{H}[{\tt P}(y_l|\bm\pi)]\big]
	\end{align}\endgroup
	The sequence-level  uncertainties, or rates, in~\cite{structured prediction} may be obtained by normalising the quantities in eqs. (\ref{eq:total approx}) and (\ref{eq:data approx}) by the sequence length.
	
	\subsection{Guided Uncertainty Approach}
	
	Training distribution distilled models can be significantly harder than training to standard distillation. Hence, a two model approach will also be explored. This is based on the observation that the distribution distilled model tends to have high Spearman's rank correlation with the ensemble predicted uncertainties in teacher-forcing mode used in training. In contrast, evaluating the model in free-run decoding, so predictions are based on ${\hat{\bm y}}_{<l}$, the same consistency between the model and the ensemble was not observed. To address this, the distribution distilled model was fed the back-history from the distilled model, as the distilled model was found to be less sensitive to the teacher-forcing and free-running mismatch.
	
	Assuming we have a distilled model $\bm\phi$ obtained from eq. (\ref{eq:seq dist}), and a distribution distilled model $\bm\lambda$ obtained from either eq. (\ref{eq: nll distdist}) or (\ref{eq: kl distdist}), one can then perform free-run decoding according to:
	\begin{align*}
	&\bm{\hat{\pi}}_l = \bm f(\bm{\hat{y}}_{<l}, \bm x; \bm\phi); \:\:\:\: \hat{y}_l\sim{\tt P}(y_l|{\hat {\bm y}}_{<l},{\bm x};{\bm\phi}) \\
	&\hat{\bm\alpha_l} = \bm f(\bm{\hat{y}}_{<l}, \bm x; \bm\lambda) 
	\end{align*}
	The distilled model $\bm\phi$ is used to predict the output sequence, which then guides the second model $\bm\lambda$ to return higher quality uncertainties that can be derived from $\{\hat{\bm\alpha_l}\}_{l = 1}^{L}$. Although this method (referred to as guided uncertainty approach; GUA) does not yield the same efficiency as standard distribution distillation, it ensures that the best attributes of $\bm\phi$ and $\bm\lambda$ are maintained in testing.

	\section{Experimental Evaluation}
	\label{sec:steup}
	
	\subsection{Data and Experimental Setup}
	
	The data used in this paper is the same as those described in \cite{edie}. This data has been manually annotated with grammatical errors and corrections. The training set and FCE (a specific test set) have been taken from the Cambridge Learner Corpus \cite{clc} and includes written exams of candidates with different L1 languages. The FCE test set is a held-out subset of the corpus and therefore, in-domain (ID) with the training data. 
	NICT-JLE \cite{nict} is a public speech corpus based on non-native speech. Only manual transcriptions of these interviews involving Japanese learners are available, along with the grammatical corrections. No audio is available. BULATS \cite{blt,DBLP:conf/icassp/KnillGMC19} is a corpus based on a free speaking business English test, where candidates were prompted for responses up to 1 minute. Both manual (BLT) and ASR (BLT$_{\text{asr}}$) transcriptions were used, the average ASR WER for this data was 19.5\% (see \cite{DBLP:conf/interspeech/LuGKMWW19} for details). The candidates are drawn from 6 L1s and span the full range of proficiencies in English. These sets are out-of-domain (OOD) and will be used to test the performance of any uncertainties derived from distribution distilled models and ensembles, see Table (\ref{table:datastats}).
	\begin{table}[h!]
		\centering
		\vspace{-1mm}
		\begin{tabular}{c|c|cccc}
			\toprule
			Set & Train & FCE & NICT & BLT & BLT$_{\text{asr}}$ \\
			\midrule
			\# of Sentences & 1.8M & 2.7K & 21.1K & 3.7K & 3.6K\\
			\midrule
			Ave. Length & 13.4 & 14.0 & 6.6 & 16.6 & 16.7 \\
			\midrule 
			Domain & Ref. & ID & OOD & OOD & OOD \\
			\bottomrule
		\end{tabular}
		\vspace{-0.2cm}
		\caption{Number of sentences, mean sentence length, and domain.}
		\label{table:datastats}
	\end{table}
	
	All models used in this work are transformers based on \cite{attention is all}, with the default parameters of 6 layers, $d_{model} = 512$, $d_{ff} = 2048$, 8 heads, and 10\% dropout. Input words were mapped to randomly initialized embeddings of dimension 512. Random seeds were used to generate an ensemble of 5 models $M = 5$ \cite{seed ensemble}. As 5 models are used in the ensemble, this increases the memory requirements by a factor 5 over a single model, and approximately 5 for the decoding cost depending on the form of beam-search being used.	It is possible to utilize more advanced ensemble generation methods, but is not key for this work \cite{back BNN, dropout ensemble, baseline BNN}. Additionally, sequence models can make predictions in different ways: \textit{expectation of products}, and \textit{products of expectations} \cite{structured prediction}. Since it has been found in \cite{structured prediction} that products of expectations performs better, it will be adopted in this work when evaluating ensembles. Beam-search with a beam of one will be used, and uncertainty metrics will be evaluated on this single output.
	
	
	
	As noted in \cite{ensemble dist dist}, when training, the target categorical is often concentrated in one of the classes, while the Dirichlet predicted by the student often has a  distribution spread over many classes. This implies that the common support between the teacher and student is poor. Optimizing KL-divergence when there is a small non-zero common support between target and prediction can lead to instability, and therefore, temperature annealing will be used to alleviate this \cite{knowledge distillation}. In this work, when training a distribution distilled system, the targets from the ensemble are annealed with a temperature $T$, following a schedule from $T=10.0$ to $3.0$. Further reducing the temperature down to $T = 1.0$ resulted in instability during training. To remain consistent, all results concerning ensembles will also be based on the final temperature $T=3.0$.
	
	
	The GEC performance metric used in this work is GLEU \cite{gleu}, and uncertainty metrics will be based at the token level. Furthermore, the distribution distilled models and ensembles will be compared when a fraction of samples are rejected (based on some metric) and passed on to an oracle. In these cases, the rejection will be based on the highest sentence level uncertainty metrics, and will be compared to simply rejecting sentences based on their length. An additional metric based on knowing the true target will be ${\tt L} \cdot (1 - \GLEU)$ (referred to as {\tt manual}), where ${\tt L}$ is the true sequence length; the inclusion of length is important as sentence level GLEU does not take length into account even though it has an effect when used at a dataset level \cite{gleu 2}. System performance will be evaluated using relative area under the curve $\AUCRR$ as defined by:
	\begin{align}
	\AUCRR = \frac{\AUC - \AUC_{\tt random}}{\AUC_{\tt manual} - \AUC_{\tt random}}
	\end{align}
	following the same reasoning as in \cite{aucrr} to simplify comparison between metrics; $\AUC_{\tt random}$ refers to fully random rejection. Finally GLEU performance will also be reported at 10\% rejection.
	
	\subsection{Results}
	\label{sec:res}
	
	Table \ref{tab:gec-models} shows the GLEU performance of a range of different models. As expected the ensemble performs best, showing performance gains over the individual ensemble members. Distilling the ensemble, using eq. (\ref{eq:seq dist}), again yielded performance gains over the individual ensemble members, though not to the same level as the original ensemble but at reduced memory and computational cost.
	\begin{table}[h!]
		\centering{}
		\begin{minipage}[t]{0.47\textwidth}%
			\begin{center}
				\begin{tabular}{c||c|c|c||c|c|c}
					\toprule
					\textbf{Test set} & {Ind}$_{\pm\sigma}$ & {Ens.} & {Dist.} & {NLL} & {KL} & {GUA}\\
					\midrule  
					FCE & 69.5$_{\pm 0.11}$                       & 70.6 & 69.9 & 68.0 & 68.8 & 69.9 \\
					NICT & 47.2$_{\pm 0.20}$                     & 48.0 & 47.8 & 44.7 & 45.7 & 47.8 \\
					BLT & 49.8$_{\pm 0.15}$                       & 50.9 & 50.7 & 48.2 & 48.9 & 50.7 \\
					BLT$_{\text{asr}}$ & 31.3$_{\pm 0.09}$ & 31.6 & 31.5 & 30.9 & 31.3 & 31.5 \\
					\bottomrule
				\end{tabular}
				\par\end{center}
		\end{minipage}
		\vspace{-1mm}
		\caption{GLEU performance of individual, ensemble and standard and distribution distilled models. Ind gives shows the ensemble member mean GLEU, and standard deviation.}
		
		\label{tab:gec-models}
	\end{table}
	Secondly, the table also shows the performance of distribution distilled models. The simplified training using KL-divergence,  eq. (\REFKL), outperforms NLL,  eq. (\REFNLL), though neither matches the average performance of the ensemble members. This illustrates the challenges in training sequence ensemble distribution distillation models. As the KL model performed better, it will be used as the uncertainty model in GUA, though both the NLL and KL based models yielded high Spearman rank correlation coefficient with the ensemble uncertainty measures. The same trend can be seen for both the in-domain data (FCE) and the out-of-domain data.
	
	Table \ref{tab:gec-uncertainty} shows the average word-level uncertainties predicted for two datasets, FCE and the most mismatched BLT$_{\text{asr}}$.
	\begin{table}[h!]
		\centering{}
		\begin{minipage}[t]{0.45\textwidth}%
			\begin{center}
				\begin{tabular}{c|c|ccc}
					\toprule
					\textbf{Test Set} & \textbf{Model} & TU & DU & KU \\
					\midrule 
					\multirow{2}{*}{FCE} & Ensemble & 8.41 & 8.24 & 0.16 \\
					& GUA & 8.37 & 8.20 & 0.18 \\
					\midrule 
					\multirow{2}{*}{BLT$_{\text{asr}}$} & Ensemble & 8.85 & 8.67 & 0.18 \\
					& GUA & 8.85 & 8.65 & 0.20 \\
					\bottomrule 
				\end{tabular}
				\vspace{-2mm}
				\caption{Word-level uncertainties for an in-domain (FCE) and out-of-domain (BLT$_{\text{asr}}$) dataset. Total (TU), Data (DU) and Knowledge (KU) Uncertainties.}
				\label{tab:gec-uncertainty}
				\par\end{center}
		\end{minipage}
	\end{table}
	The guided model approach behaves similarly to the ensemble. High knowledge uncertainty (KU) is, according to theory, an indication of a sample being OOD and in this case as expected KU is higher for BLT$_{\text{asr}}$—an out of-of-domain test set. Furthermore, it can be seen that all uncertainties for BLT$_{\text{asr}}$ are higher than FCE across both models. 
	
	Table \ref{tab:gec-relative auc} shows relative AUC (AUC$_{\tt RR}$) for FCE and BLT$_{\text{asr}}$, as well as mix of the two which assess whether the uncertainty measures can detect the more challenging OOD speech data.
	It is interesting that as a baseline simply using the length of the of the output sequence (Length) is a good baseline, as longer sentences will tend to have more complex grammatical structure and opportunity for mistakes. KU performs best in rejecting challenging inputs. for both the ensemble and GUA, outperforming simple length selection for all conditions. 
	\begin{table}[h!]
		\centering{}
		\makebox[0.45\textwidth][c]{
			\resizebox{1.0\textwidth}{!}{
				\begin{minipage}[t]{1.0\textwidth}%
					\begin{center}
						\begin{tabular}{c|c|cccc}
							\toprule
							\textbf{Test Set} & \textbf{Model} & Length & TU & DU & KU \\
							\midrule 
							\multirow{2}{*}{FCE} & Ensemble & 0.701 & 0.734 & 0.734 & \textbf{0.740} \\
							& GUA & 0.701 & 0.736 & 0.735 & \textbf{0.750} \\
							\midrule 
							\multirow{2}{*}{BLT$_{\text{asr}}$} & Ensemble & 0.895 & \textbf{0.914} & 0.914 & 0.909 \\
							& GUA & 0.895 & 0.902 & 0.901 & \textbf{0.917} \\
							\midrule
							\midrule
							FCE +  & Ensemble & 0.810 & 0.840 & 0.840 & \textbf{0.843} \\
							BLT$_{\text{asr}}$ & GUA & 0.813 & 0.837 & 0.837 & \textbf{0.856} \\
							\bottomrule 
						\end{tabular}
						\par\end{center}
				\end{minipage}
		}}
		\vspace{-2mm}
		\caption{AUC$_{\tt RR}$ performance. The combined FCE + BLT$_{\text{asr}}$ refers to the case when the system is run on both ID and OOD data.}
		\label{tab:gec-relative auc}
	\end{table}
	
	Table \ref{tab:gec-rejection} shows the performance when a fixed percentage, 10\%, is "rejected" and manually corrected. These results can be compared to the baseline numbers in Table \ref{tab:gec-models}, show significant gains from rejecting 10\% of the data. Similar trends to that shown in AUC$_{\tt RR}$ can be seen, KU generally yields the best performance, outperforming the baseline length approach in all conditions. GUA yields similar performance to the original ensemble, but at reduced memory and computational costs.
	
	\begin{table}[h!]
		\centering{}
		\makebox[0.40\textwidth][c]{
			\resizebox{1.0\textwidth}{!}{
				\begin{minipage}[t]{1.0\textwidth}%
					\begin{center}
						\begin{tabular}{c|c|cccc|c}
							\toprule
							\textbf{Test Set} & \textbf{Model} & Length & TU & DU & KU & Manual\\
							\midrule 
							\multirow{2}{*}{FCE} & Ens. &  80.5 & 80.8 & 80.8 & \textbf{81.0} & 83.4 \\
							& GUA & 80.1 & 80.5 & 80.5 & \textbf{80.5} & 82.9 \\
							\midrule 
							\multirow{2}{*}{BLT$_{\text{asr}}$} & Ens. & 52.4 & \textbf{54.4} & 54.2 & 53.2 & 56.5 \\
							& GUA & 51.9 & 52.6 & 52.5 & \textbf{53.4} & 56.3 \\
							\midrule 
							\midrule
							FCE + & Ens. & 65.9 & 66.9 & 66.8 & \textbf{66.9} & 68.7 \\
							BLT$_{\text{asr}}$ & GUA & 65.3 & 65.8 & 65.8 & \textbf{66.2} & 68.3 \\
							\bottomrule 
						\end{tabular}
						\par\end{center}
		\end{minipage}}}
		\vspace{-2mm}
		\caption{GLEU performance when 10\% of inputs are rejected (based on a metric), and passed on to manual correction.}
		\label{tab:gec-rejection}
	\end{table}
	
	\section{Conclusion}
	\label{sec:conclusion}
	
	This work describes the application of ensemble distillation approaches to sequence data. Two forms of distillation are discussed: standard distillation where the distilled model predicts the ensemble mean; and distribution distillation where the distribution over the ensemble predictions is modelled. Though more challenging to train, ensemble distribution distillation yields both an ensemble prediction and uncertainty measures associated with prediction which allows, for example, more challenging predictions to be manually corrected. The approaches were evaluated on grammatical error correction tasks using data either matched to the written training data, or mismatched speech data. Standard distillation was found to work well for sequence distillation, but distribution distillation acting alone did not yield good performance. By the combining the predictions from standard distillation with the uncertainty predictions of distribution distillation, both good performance and uncertainty measures could be obtained, with reduced computational memory costs compared to the original ensemble. 
	
	Future work will examine: improved optimisation approaches and approximations for sequence ensemble distribution distillation; and alternative uncertainty criteria described in~\cite{structured prediction}.

	\vfill\pagebreak
	


\begin{thebibliography}{50}
		
		\providecommand{\newblock}{\relax}
		
		\bibitem{knowledge distillation}
		G. Hinton, O. Vinyals, \& J. Dean, "Distilling the Knowledge in a Neural Network," \textit{Proc. NIPS}, Montreal, Canada, 2014.
		
		\bibitem{dist object detection}
		G. Chen, W. Choi; X. Yu, T. Han, \& M. Chandraker, "Learning Efficient Object Detection Models with Knowledge Distillation, " \textit{Proc. NeurIPS}, Long Beach, CA, United States, 2017.	
		
		\bibitem{nlp kd 1}
		J. Cui, B. Kingsbury, B. Ramabhadran, G. Saon, T. Sercu, K. Audhkhasi, A. Sethy, M. Nussbaum-Thom, \& A. Rosenberg, "Knowledge Distillation Across Ensembles of Multilingual Models for Low-resource Languages," \textit{Proc. ICASSP}, New Orleans, LA, United States, 2017.
		
		\bibitem{nlp kd 2}
		R. Yu, A. Li, V. I. Morariu, \& L. S. Davis, "Visual Relationship Detection with Internal and External Linguistic Knowledge Distillation," \textit{Proc. ICCV}, Venice, Italy, 2017.
		
		\bibitem{DBLP:conf/interspeech/WongG16}
		J. H. M. Wong \& M. J. F. Gales, "Sequence Student-teacher Training of Deep Neural Networks,'' \textit{Proc. Interspeech}, San Francisco, CA, United States, 2016.
		
		\bibitem{dist lm}
		T. Asami, R. Masumura, Y. Yamaguchi, H. Masataki, \& Y. Aono, "Domain Adaptation of DNN Acoustic Models using Knowledge Distillation," \textit{Proc. ICASSP}, New Orleans, LA, United States, 2017.
	
		\bibitem{end for adverserial attack}
		N. Papernot, P. McDaniel, X. Wu, S. Jha, \& A. Swami, “Distillation as a Defense to Adversarial Perturbations Against Deep Neural Networks," \textit{Proc. SP}, San José, Costa Rica, 2016.
		
		\bibitem{murphy}
		Kevin P. Murphy. \textit{Machine Learning}. The MIT Press, 2012.
		
		\bibitem{seed ensemble}
		B. Lakshminarayanan, A. Pritzel, \& C. Blundell, "Simple and Scalable Predictive Uncertainty Estimation using Deep Ensembles,"  \textit{Proc. NeurIPS}, Long Beach, CA, United States, 2017.
		
		\bibitem{prior net}
		A. Malinin \& M. J. F. Gales, "Predictive Uncertainty Estimation via Prior Networks," \textit{Proc. NeurIPS}, Montreal, Canada, 2018.
		
		\bibitem{ensemble dist dist}
		A. Malininm, B. Mlodozeniec, \& M. J. F. Gales, "Ensemble Distribution Distillation," \textit{Proc. ICLR}, 2020.
		
		\bibitem{ensemble uncertanity spoken}
		X. Wu , K. M. Knill, M. J. F. Gales, \& A. Malinin, “Ensemble Approaches for Uncertainty in Spoken Languagege Assessment," \textit{Proc. Interspeech}, Shanghai, China, 2020.
		
		\bibitem{seq endd}
		Y. Kim \& A. M. Rush, “Sequence-Level Knowledge Distillation," \textit{Proc. EMNLP}, Austin, TX, United States, 2016.
		
		\bibitem{structured prediction}
		A. Malinin \& M. J. F. Gales, "Uncertainty in Structured Prediction," \textit{arXiv.org}, 2002.07650v3, 2020.
		
		\bibitem{understanding kd}
		C. Zhou, J. Gu, \& G. Neubig, "Understanding Knowledge Distillation in Non-autoregressive Machine Translation," \textit{Proc. ICLR}, 2020.
		
		\bibitem{gec dnn task}
		Z. Yuan \& T. Briscoe, “Grammatical Error Correction Using Neural Machine Translation," \textit{Proc. ACL}, San Diego, CA, United States, 2016.
		
		\bibitem{compression e2e}
		R. Pang, T. N. Sainath, R. Prabhavalkar, S. Gupta, Y. Wu, S. Zhang, \& C. Chiu, "Compression of End-to-End Models", \textit{Proc. Interspeech}, Hyderabad, India, 2018.
		
		\bibitem{back BNN}
		J. M. Hernandez-Lobato \& R. Adams, "Probabilistic Backpropagation for Scalable Learning of Bayesian Neural Networks," \textit{Proc. ICML}, Lille, France, 2015.
		
		\vfill\pagebreak
		
		\bibitem{dropout ensemble}
		Y. Gal \& Z. Ghahramani, "Dropout as a Bayesian Approximation: Representing Model Uncertainty in Deep Learning," \textit{Proc. ICML}, New York City, NY, United States, 2016.
		
		\bibitem{adverserial uncertainty}
		L. Smith  \& Y. Gal, “Understanding Measures of Uncertainty for Adversarial Example Detection,” \textit{Proc. UAI}, Monterey, CA, United States, 2018.
		
		\bibitem{ygal uncertainty}
		Y. Gal, "Uncertainty in Deep Learning", Ph.D. thesis, University of Cambridge, 2016.
		
		\bibitem{dataset shift}
		J. Quiñonero-Candela, "Dataset Shift in Machine Learning", \textit{The MIT Press}, 2009.
		
		\bibitem{more seq kd}
		M. Huang, Y. You, Z. Chen, Y. Qian, \& K. Yu, "Knowledge Distillation for Sequence Model," \textit{Proc. Interspeech}, Hyderabad, India, 2018.
		
		\bibitem{edie}
		Y. Lu, M. J. F. Gales, \& Y. Wang, “Spoken Language ‘Grammatical Error Correction’," \textit{Proc. Interspeech}, Shanghai, China, 2020.
		
		\bibitem{clc}
		D. Nicholls, “The Cambridge Learner Corpus: Error Coding and Analysis for Lexicography and ELT,” \textit{Proc. Corpus Linguistics}, Lancaster, United Kingdom, 2003.
		
		\bibitem{nict}
		E. Izumi, K. Uchimoto, \& H. Isahara, “The NICT JLE Corpus Exploiting the Language Learners' Speech Database for Research and Education,” \textit{Proc. IJCIM}, Bangkok, Thailand, 2004.
		
		\bibitem{blt}
		L. Chambers \& K. Ingham, “The BULATS Online Speaking Test,” \textit{Research Notes}, vol. 43, pp. 21–25, 2011. [Online]. Available: http://www.cambridgeenglish.org/images/23161-research-notes-43.pdf
		
		\bibitem{DBLP:conf/icassp/KnillGMC19}
		K. M. Knill, M. J. F. Gales, P. P. Manakul, \& A. Caines, "Automatic Grammatical Error Detection of Non-native Spoken Learner English,'' \textit{Proc. ICASSP}, Brighton, United Kingdom, 2019.
		
		\bibitem{DBLP:conf/interspeech/LuGKMWW19}
		Y. Lu, M. J. F. Gales, K. M. Knill, P. P. Manakul, L. Wang, \& Y. Wang, "Impact of {ASR} performance on spoken grammatical error detection,''  \textit{Proc. Interspeech}, Graz, Austria, 2019.
		
		\bibitem{attention is all}
		A. Vaswani, N. Shazeer, N. Parmar, J. Uszkoreit, L. Jones, A. N. Gomez, \& Ł. Kaiser, "Attention is All You Need," \textit{Proc. NeurIPS}, Long Beach, CA, United States, 2017.
		
		\bibitem{baseline BNN}
		W. Maddox, T. Garipov, P. Izmailov, D. P. Vetrov, \& A. G. Wilson, "A Simple Baseline for Bayesian Uncertainty in Deep Learning," \textit{Proc. NIPS}, Vancouver, Canada, 2019.
		
		\bibitem{gleu}
		C. Napoles, K. Sakaguchi, M. Post, \& J. Tetreault, "Ground truth for grammatical error correction metrics," \textit{Proc. ACL}, Beijing, China, 2015.
		
		\bibitem{gleu 2}
		Y. Wu, M. Schuster, Z. Chen, Q. V. Le, et al. "Google’s Neural Machine Translation System: Bridging the Gap between Human and Machine Translation," \textit{arXiv.org}, 1609.08144v2, 2016.
		
		\bibitem{aucrr}
		A. Malinin, A. Ragni, K. M. Knill, \& M. J. F. Gales, "Incorporating Uncertainty into Deep Learning for Spoken Language Assessment, " \textit{Proc. ACL}, Vancouver, Canada, 2017.
		
	\end{thebibliography}
\end{document}